\documentclass{article}

\usepackage{url}
\usepackage{microtype}
\usepackage{graphicx}
\usepackage{subfigure}
\usepackage{booktabs} % for 
\usepackage{tikz}
\usetikzlibrary{positioning,arrows.meta,fit,shapes.geometric}
\usepackage{hyperref}

\usepackage[accepted]{icml2025}

\hypersetup{pdftitle={When Does GEPA Merge Help? A Mechanism-Level Diagnostic},
  pdfauthor={Chong Zhao, Ruyi Yang, Changming Liu, Alice Zhao},
  pdfsubject={Preprint}}
\icmltitlerunning{When Does GEPA Merge Help? A Mechanism-Level Diagnostic}

\usepackage{amsmath}
\usepackage{amssymb}
\usepackage{mathtools}
\usepackage{amsthm}

\usepackage[capitalize,noabbrev]{cleveref}

\theoremstyle{plain}

\theoremstyle{definition}

\theoremstyle{remark}

\usepackage[disable,textsize=tiny]{todonotes}

\begin{document}

\twocolumn[
\icmltitle{When Does GEPA Merge Help? A Mechanism-Level Diagnostic}
\icmlsetsymbol{equal}{*}

\begin{icmlauthorlist}
\icmlauthor{Chong Zhao}{harvard}
\icmlauthor{Ruyi Yang}{harvard}
\icmlauthor{Changming Liu}{harvard}
\icmlauthor{Alice Zhao}{harvard}

%\icmlauthor{}{sch}
%\icmlauthor{}{sch}
%\icmlauthor{}{sch}
\end{icmlauthorlist}

\icmlaffiliation{harvard}{Harvard University, Cambridge, MA, USA}

% You may provide any keywords that you
% find helpful for describing your paper; these are used to populate
% the "keywords" metadata in the PDF but will not be shown in the document
% \icmlkeywords{Machine Learning, ICML}

\vskip 0.3in
]

% Affiliation footnote (preprint; no conference notice)
{\let\thefootnote\relax\footnotetext{\textsuperscript{1}Harvard University, Cambridge, MA, USA.}}

\begin{abstract}
GEPA is among the strongest reflective prompt optimizers for language-model programs, but its system-aware merge operator is unstable across models: merge improves gpt-4.1-mini on average while degrading Qwen3-8B, an asymmetry noted but left unexplained by the GEPA paper. We investigate this instability across five datasets and two models. Our nine-configuration sweep over merge timing and merge operator shows that merge can improve both models when configured appropriately. This suggests that merge instability is better explained by configuration sensitivity than by an inherent flaw in the merge operator. We then test whether prompt-level features predict GEPA’s merge-accept decisions and find that they do not, ruling out static-feature gating. As a proof of concept, a behavior-based probe built from these diagnostics avoids large merge failures in our multi-seed runs by abstaining on high-risk tasks, although it does not consistently outperform NoMerge. Code is available as GEPA-MergeProbe at \url{https://github.com/ChongZZZZ/GEPA_MERGE}.

\end{abstract}

\section{Introduction}
\label{sec:intro}
Prompt optimization without parameter updates has emerged as a central technique for adapting language models to new tasks, and evolutionary search has become competitive with reinforcement learning at a fraction of the rollout cost, offering a parameter-free alternative to fine-tuning~\cite{agrawal2026gepa}. GEPA is one of the strongest such methods. It maintains a Pareto frontier of candidates and iteratively applies two operators: reflective mutation, which rewrites instructions based on model self-critique, and a crossover-style merge that synthesizes a new candidate from two evolved parents sharing a common ancestor.

\begin{figure}[ht]
\centering
\includegraphics[width=\columnwidth]{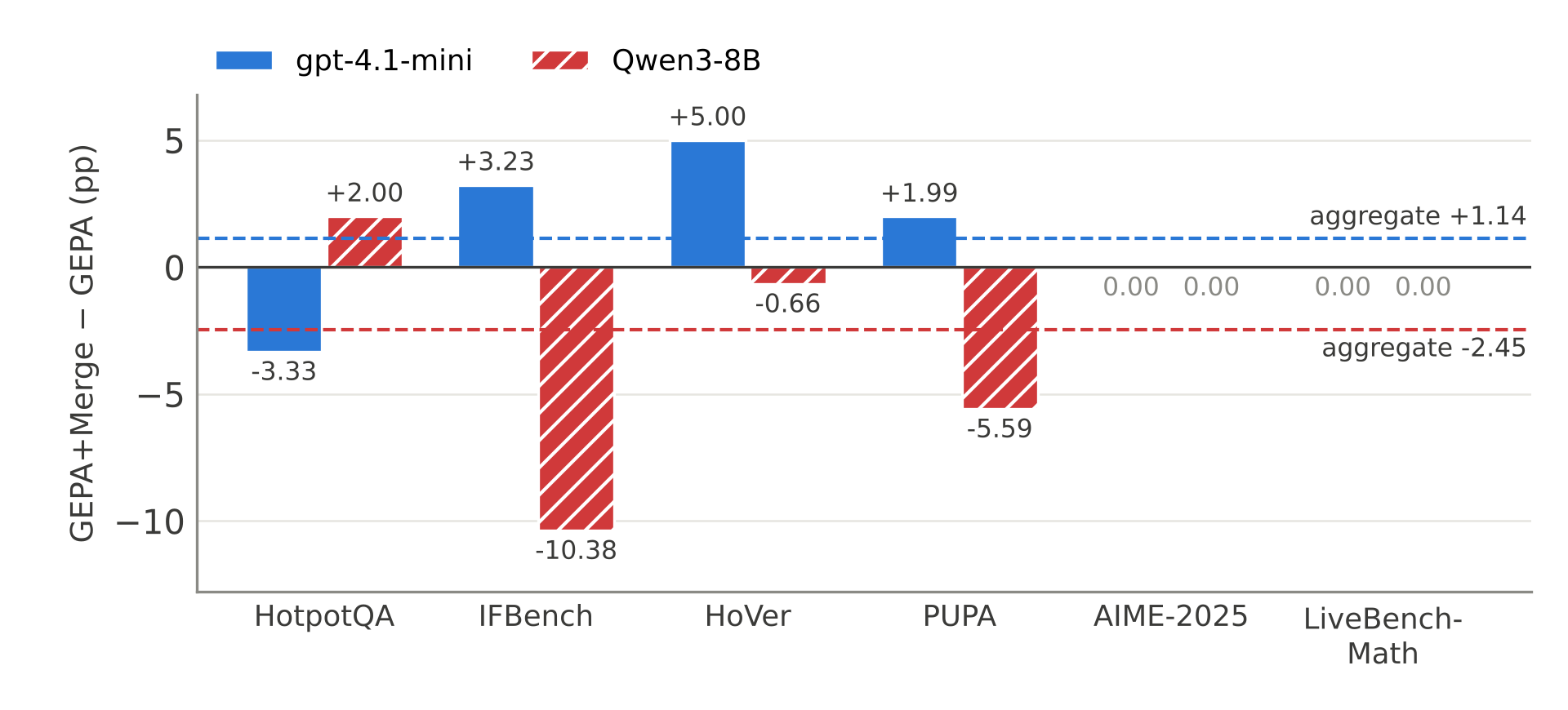}
\caption{Test-score change from adding merge to GEPA (GEPA+Merge $-$ GEPA), computed from Tables~1--2 of \citet{agrawal2026gepa}. Merge helps gpt-4.1-mini but hurts Qwen3-8B; dashed lines mark the six-task aggregate change ($+1.14$ and $-2.45$~pp).}
\label{fig:obs5}
\end{figure}

Merge behavior, however, is unstable across models and datasets. In the results reported by \citet{agrawal2026gepa}, using an identical configuration, merge improves gpt-4.1-mini by $+1.14$~pp on average while degrading Qwen3-8B by $-2.45$~pp (Figure~\ref{fig:obs5}). The original paper labels this asymmetry and leaves the underlying mechanism unexplained.

The GEPA paper attributes this asymmetry to fixed merge parameters, such as the timing schedule and number of merges. These parameters are shared across model scales rather than adapted to each model’s optimization dynamics, and the paper argues that they should be adaptive. However, the paper does not ablate either one. We perform these ablations directly.

We ask whether the instability is algorithmic, meaning intrinsic to the merge operator itself, or hyperparametric, meaning fixable by adjusting when and how merge is invoked. We answer this question through a configuration sweep, a diagnostic analysis of merge outcomes, and a behavior-based probe that operationalizes the resulting mechanisms.

\paragraph{Contributions.}
\begin{itemize}\setlength{\itemsep}{2pt}
\item \textbf{A broad characterization sweep of GEPA merge}, spanning 5~datasets $\times$ 2~models $\times$ 9~(timing $\times$ operator) configurations against NoMerge baselines.

\item \textbf{A mechanism-level diagnostic suite} showing that prompt-level features cannot predict merge acceptance, reflection provides substantially larger direct validation gains than merge, and val/test correlation bounds task-level selection reliability.

\item \textbf{A multi-seed variance audit} showing that NoMerge itself has several-pp seed variation, which sets a noise floor for interpreting single-seed merge deltas.

\item \textbf{A proof-of-concept behavioral probe} that operationalizes these diagnostics and avoids large tail failures in multi-seed runs, without consistently outperforming NoMerge.
\end{itemize}

\section{Related Work}
\label{sec:related_work}
Our question sits at the intersection of two lines of work: evolutionary search over language-model artifacts and reflective optimization of multi-module LM programs. The first motivates GEPA+Merge as a promising population-based optimizer. The second shows that optimization-loop structure, which includes reflection, parent selection, and merge, can determine whether LM-program search converges or destabilizes, making merge behavior a natural target for mechanistic analysis.

\paragraph{Evolutionary search over prompts and programs.} Population-based optimization of natural-language artifacts has emerged as a practical alternative to parameter fine-tuning. EvoPrompt~\cite{guo2024evoprompt} casts prompt optimization as a discrete evolutionary process with LM-driven variation operators and outperforms human-engineered prompts on several benchmarks, including BIG-Bench Hard. 
Parallel threads extend the paradigm to adversarial prompt generation via quality-diversity search (Rainbow Teaming~\cite{samvelyan2024rainbow}) and to program synthesis (AlphaEvolve~\cite{novikov2025alphaevolve}). Together, these works establish that evolutionary search over LM-generated artifacts is broadly effective.

\paragraph{Reflective prompt optimization for LM programs.} A complementary line targets LM programs: multi-module pipelines whose modules each receive their own instruction. MIPROv2~\cite{opsahl-ong-etal-2024-optimizing} jointly optimizes instructions and few-shot demonstrations via Bayesian candidate proposal. GEPA~\cite{agrawal2026gepa} introduces reflective mutations, in which an LM critiques its own incorrect outputs and rewrites the offending instruction, paired with a system-aware merge operator that combines two evolved candidates sharing a common ancestor. GEPA matches or exceeds reinforcement-learning fine-tuning at up to $35\times$ fewer rollouts but leaves the question of when to invoke merge explicitly unresolved. We address that question through a systematic diagnostic.

\section{Setup: GEPA Merge and Evaluation Protocol}
\label{sec:background}
\subsection{GEPA pipeline}
\label{sec:setup-pipeline}
GEPA~\cite{agrawal2026gepa} optimizes prompts for LM programs: multi-module pipelines in which each module receives its own natural-language instruction. It does so by iteratively proposing and evaluating new prompt candidates under a fixed evaluation budget. 
\begin{figure*}[t]
    \centering
    \includegraphics[width=0.9\textwidth]{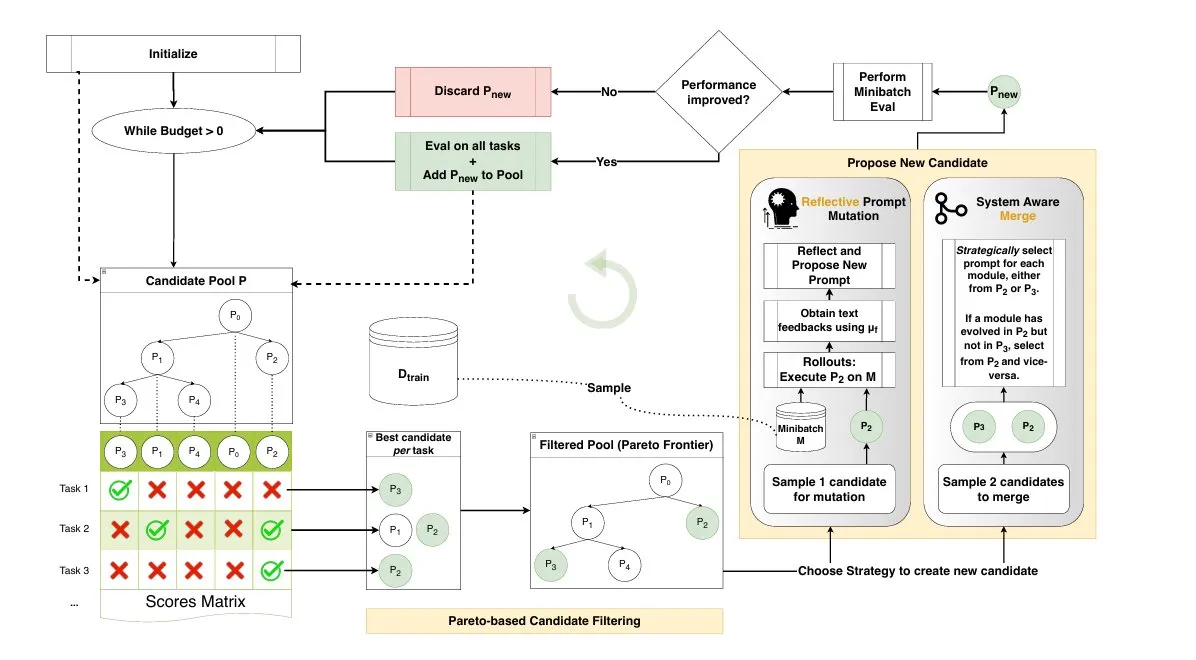}
    \caption{Overview of the GEPA optimization loop, reproduced from \citet{agrawal2026gepa}. Parents are drawn from a Pareto-filtered candidate pool (bottom); each iteration proposes a child with either reflective prompt mutation or system-aware merge (right), and keeps it only if it improves on a minibatch and then on the full validation set (top). Our study varies \emph{when} the merge branch may fire and \emph{how} it resolves conflicting modules (\S\ref{sec:design-space}).}
    \label{fig:gepa-overview}
\end{figure*}

As shown in Figure~\ref{fig:gepa-overview}, at each iteration, GEPA samples a parent from a Pareto-frontier-organized candidate pool, applies one of two operators to produce a child, evaluates the child on a minibatch, and re-evaluates on the full validation set if the minibatch shows improvement. Accepted candidates re-enter the pool. The two operators are reflective mutation, in which an LM critiques the parent's incorrect outputs and rewrites the instruction, and system-aware merge, in which two evolved candidates sharing a common ancestor are combined into a new candidate. Because our analysis focuses on merge, we next describe the merge operator in detail.

\subsection{Merge operator}
\label{sec:setup-merge}
For each predictor in the LM program, GEPA's merge operator (Algorithm~4 of \cite{agrawal2026gepa}) takes a triplet $(\mathit{anc}, p_1, p_2)$ consisting of a common ancestor and two descendants. It then classifies the descendants' divergence from the ancestor into three cases, summarized in Table~\ref{tab:merge-cases}.

A valid merge must pass three gates: an ancestry validity check requiring a common ancestor, a desirable-predictor filter requiring the merged predictor to differ from the ancestor on at least one component, and a behavioral subsample accept gate requiring the merged candidate to strictly improve over both parents on a validation subsample. The final gate is important because GEPA does not accept a merge based only on how reasonable the merged prompt looks. Instead, it accepts the merge only if the merged candidate performs better than both parents on a validation subsample.

\begin{table}[ht]
\centering
\caption{Case classification in the GEPA merge operator.}
\label{tab:merge-cases}
\small
\begin{tabular}{@{}p{0.5cm}p{3cm}p{3cm}@{}}
\toprule
\textbf{Case} & \textbf{Condition} & \textbf{Resolution} \\
\midrule
A & Exactly one descendant diverged from ancestor & Take the diverged descendant \\
B & Both diverged differently & Variant-dependent (\S\ref{sec:design-space}) \\
C & Both descendants agree ($p_1 = p_2$) & Take either \\
\bottomrule
\end{tabular}
\end{table}

\subsection{Benchmarks and models}
\label{sec:setup-benchmarks}
We evaluate on five benchmarks (Table~\ref{tab:benchmarks}). Three datasets---HotpotQA \cite{yang-etal-2018-hotpotqa}, HoVer \cite{jiang-etal-2020-hover}, and IFBench \cite{pyatkin2025ifbench}---are retained from the GEPA paper to preserve comparability. We exclude PUPA due to Qwen3-8B's low parse rate in a NoMerge sanity check. We add MuSiQue \cite{trivedi-etal-2022-musique} and 2WikiMultiHopQA \cite{ho2020constructing} as additional multi-hop probes. These multi-hop tasks target a setting where merge may be especially useful: different parent prompts can specialize in different reasoning or retrieval trajectories, creating opportunities for merge to combine complementary behaviors.

Our implementation builds on DSPy~\citep{khattab2023dspy}, matching
GEPA's implementation framework. We use gpt-4.1-mini and Qwen3-8B as both the program LM and the corresponding reflector, matching the GEPA paper’s cross-model setup. For computational reasons, Qwen3-8B is run with thinking mode disabled. Each held-out test evaluation uses 300 examples, and we use the same per-task metric-call budgets as the GEPA paper, ranging from $3{,}593$ on IFBench to $7{,}051$ on HoVer. We also use the same base prompts as the GEPA paper.

\begin{table}[ht]
\centering
\caption{The five benchmarks evaluated. Per-benchmark retrieval modes are reported in Appendix~\ref{app:retrieval-details}.}
\label{tab:benchmarks}
\small
\begin{tabular}{@{}p{2.6cm}p{3.6cm}@{}}
\toprule
\textbf{Dataset} & \textbf{Task type} \\
\midrule
HotpotQA          & multi-hop QA \\
HoVer             & fact verification \\
IFBench           & instruction following \\
MuSiQue           & compositional multi-hop QA \\
2WikiMultiHopQA   & entity-relation multi-hop QA \\
\bottomrule
\end{tabular}
\end{table}

\section{Configuration Sweep: Is Merge Broken or Misconfigured?}
\label{sec:design-space}
\subsection{Timing and operator design space}
\label{sec:design-space-axes}

The instability in Figure~\ref{fig:obs5} suggests that merge may not be universally good or bad, but it is sensitive to how it is invoked. In GEPA's default implementation, two choices are fixed: merge is allowed to fire immediately after the first successful reflective mutation, and disputed
predictors are resolved using GEPA's original per-predictor rule. If the failure is due to choice of hyperparameters, then changing these choices should change merge outcomes.

We therefore ablate two dimensions of merge behavior: \emph{when} merge is allowed to fire and \emph{how} conflicting parent instructions are combined. Three timing policies and three operator variants form a $3\!\times\!3$ design space, evaluated against a NoMerge baseline across two models and five benchmarks. This lets us test not only whether merge can outperform NoMerge, but also under which configurations and task settings it does so. The paper-default cell (original $\times$ immediate) is one point in this space.

\paragraph{Timing policies}
The timing axis controls when merge is allowed to fire during optimization. We compare three policies:
\begin{itemize}
\item \textbf{Immediate} (GEPA's default): merge may fire from the first successful reflective mutation onward, as long as a valid pair of parents exists. We expect this policy to expose merge to the least mature parents and therefore the greatest downside risk.

\item \textbf{Score plateau} ($k\!=\!3$): merge is triggered only after $k$ consecutive reflective-mutation rounds fail to improve validation performance. We expect parents to have more time to mature before being combined.

\item \textbf{Budget proportional} ($p\!=\!0.33$): merge is withheld until fraction $p$ of the metric-call budget has been consumed, normalizing the delay across benchmarks with different budgets. We expect this to be the most conservative policy, combining only relatively mature lineages.
\end{itemize}

\paragraph{Operator variants}
The operator axis controls how merge combines parent instructions when both parents propose different changes to the same predictor. We compare three approaches: a shuffler, an accumulator, and a rewriter.
\begin{itemize}
\item \textbf{Original} (GEPA's default): selects the higher-scoring parent's instruction. This acts as a shuffler: it preserves parent content verbatim and introduces no new wording.

\item \textbf{Combine all}: concatenates both instructions in score order, preserving both parents' trajectories. This acts as an accumulator: it preserves both parents' content but increases prompt length.

\item \textbf{Summarize before}: uses an auxiliary LM to synthesize a single instruction from both parents, falling back to concatenation on failure. This acts as a rewriter and is the only variant that can introduce wording not present in either parent.
\end{itemize}

\paragraph{Outcome definitions.}
We analyze each sweep cell, corresponding to one (model, benchmark, operator, timing) configuration, by its held-out test score relative to the matched NoMerge baseline. A \textit{winner} is the merge configuration with the highest held-out test score for a given (model, benchmark) pair. A \textit{catastrophic cell} is one whose test score falls at least $20$~pp below its corresponding NoMerge baseline. These definitions determine the bolded winners and dark-red failure cells in Figure~\ref{fig:heatmap}.

\subsection{Main sweep result}
\label{sec:sweep}
\begin{figure*}[ht]
\centering
\includegraphics[width=\textwidth]{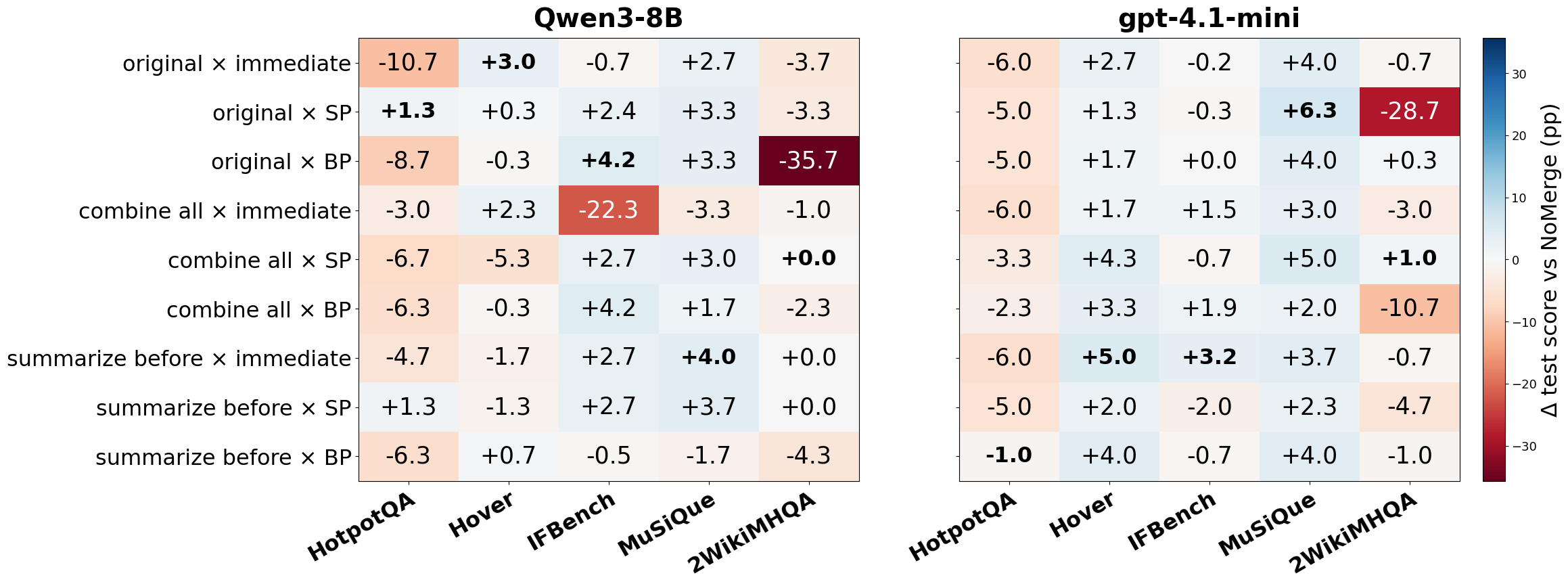}
\caption{\textbf{Configuration sweep: $\Delta$ test score relative to NoMerge.} Columns denote benchmarks and rows denote $(\text{operator}\!\times\!\text{timing})$ configurations. The left panel shows Qwen3-8B; the right panel shows gpt-4.1-mini. Per-benchmark winners are shown in boldface, and catastrophic cells with $\Delta\!\leq\!-20$~pp appear as dark-red outliers. The paper-default setting (original$\times$immediate) is rarely the best
configuration.}
\label{fig:heatmap}
\end{figure*}
Figure~\ref{fig:heatmap} reports the test-score change relative to NoMerge for every combination in the sweep. Three observations stand out. First, merge can improve over NoMerge when the timing and operator are chosen appropriately: for every $(\text{model}, \text{benchmark})$ pair, at least one merge configuration outperforms the NoMerge baseline. Second, these wins do not come from a single universally best setting. Six distinct $(\text{operator}, \text{timing})$ configurations claim at least one win, while the paper-default setting (original $\times$ immediate) is rarely among them. This supports the hyperparametric reading of GEPA merge's instability (\S\ref{sec:design-space}). Merge is useful, but only under the right configuration. Third, three strongly negative cells ($\Delta \leq -20$~pp) remain, showing that merge is conditionally useful but still unstable when timing or operator choices are poorly matched to the task. We analyze these failure cases below.

\paragraph{Pattern-level findings.}
Three patterns hold across the grid:
\begin{itemize}
\item \textbf{Operator $\times$ model.} combine\_all is the riskiest operator on Qwen3-8B: one of the three catastrophic cells uses this operator. In contrast, the same operator is benign on gpt-4.1-mini. This suggests that \textsc{combine-all} carries a prompt-bloat risk on lower-capacity models, since it increases prompt length by about $60\%$ per merge on average.

\item \textbf{Timing $\times$ lift vs.\ tail risk.}
The timing policies trade off upside and downside differently. Comparing timing choices within each fixed (model, benchmark, operator) group, score-plateau is the most consistent middle ground: it is best in 16/30 comparisons, achieves the highest single lift ($+6.34$~pp), and has the fewest catastrophes. Immediate has the largest mean lift among its winning comparisons ($+3.04$~pp), but also the highest catastrophe density, suggesting that early merge exposes the search to immature parents. Budget-proportional is more conservative in upside, with the smallest peak lift ($+4.25$~pp), but it still contains the worst tail event ($-35.67$~pp). This suggests that tail failures are not controlled by timing alone, but also depend on the operator, task, and model.

\item \textbf{Dataset feedback granularity.}
MuSiQue shows modest improvements on both models, whereas 2WikiMultiHopQA produces the largest catastrophic regressions. This pattern is consistent with their feedback granularity: MuSiQue provides richer optimization feedback, while 2WikiMultiHopQA gives sparser feedback, making bad merge decisions harder to detect. IFBench is different: its strict-format evaluation makes it especially sensitive to prompt bloat.

\end{itemize}

\paragraph{Seed dependence of winners.}
The best $(\text{operator}, \text{timing})$ setting can change across seeds. In the Qwen3-8B reruns, several tasks select different winners at seeds $1$ and $2$. We therefore treat the heatmap as evidence that merge can help under some configuration, not as proof that a particular configuration is seed-stably optimal. The only fixed merge configuration we analyze across seeds is the behavioral probe in \S\ref{sec:probe-observations}.

\subsection{Catastrophic failure cases}
\label{sec:catastrophes}

Three cells in the single-seed sweep meet our catastrophic criterion ($\Delta \leq -20$~pp). These failures fall into two interpretable patterns. The IFBench catastrophe is a \emph{prompt-bloat} failure. When combine all concatenates both parent instructions, the resulting prompt can become longer, more
redundant, and harder for a smaller model to follow, especially under strict-format evaluation. The two 2WikiMultiHopQA catastrophes instead reflect \emph{predictor-maturity imbalance}, because LM programs contain multiple predictors, two parents may be mature on different modules. A local merge can therefore combine a strong instruction in one module with an immature or ancestor-like instruction in another, breaking coordination across the pipeline.

We treat these cells as examples of failure modes, not as deterministic failures of their corresponding configurations. As the multi-seed audit in \S\ref{sec:variance-audit} shows, some catastrophic cells do not reproduce under different seeds. These failure modes motivate the diagnostic probes in \S\ref{sec:diagnostic} and the behavioral gates in \S\ref{sec:probe}.

\subsection{Multi-seed variance audit}
\label{sec:variance-audit}

The single-seed sweep treats each test score as a point estimate, but GEPA's search is stochastic: minibatch sampling, reflective proposals, and Pareto-front sampling all depend on the random seed. We therefore
audit the noise floor by rerunning the NoMerge baseline at seeds $1$ and $2$, alongside the original seed-$0$ runs.
\begin{table}[ht]
\centering
\caption{Per-task NoMerge test scores across $3$ seeds, showing the seed-variance noise floor for interpreting merge-cell deltas in Figure~\ref{fig:heatmap}. $\dagger$ has only $n\!=\!2$ seeds.}
\label{tab:variance}
\small
\begin{tabular}{@{}p{1.6cm}r r r r@{}}
\toprule
\textbf{Task} & \multicolumn{2}{c}{\textbf{GPT-4.1-mini}} & \multicolumn{2}{c}{\textbf{Qwen3-8B}} \\
\cmidrule(lr){2-3}\cmidrule(lr){4-5}
              & \textbf{mean} & \textbf{$\sigma$} & \textbf{mean} & \textbf{$\sigma$} \\
\midrule
HotpotQA & $58.2$ & $4.0$ & $51.9$ & $4.0$ \\
IFBench  & $49.6$ & $0.8$ & $34.4$ & $2.2$ \\
HoVer    & $46.2$ & $2.1$ & $44.4$ & $3.0$ \\
MuSiQue  & $48.1$ & $4.7$ & $47.2$ & $5.1$ \\
2WikiMHQA & $83.5^{\dagger}$ & $0.2^{\dagger}$ & $82.6$ & $2.2$ \\
\bottomrule
\end{tabular}
\end{table}

\textbf{NoMerge itself is seed-sensitive.}
As shown in Table~\ref{tab:variance}, NoMerge test scores move by several percentage points across seeds. The standard deviation is about $3$--$5$~pp on the noisier tasks, especially HotpotQA and MuSiQue. This
means that seed variance is a baseline property of GEPA's stochastic search, not a phenomenon introduced only by merge.

This noise floor changes how we interpret the sweep. First, individual single-seed cells should be read cautiously: small positive or negative $\Delta$ values may fall within the NoMerge noise floor. Second, the
catastrophic cells in \S\ref{sec:catastrophes} are better viewed as rare-but-large tail events of a noisy search process, rather than as deterministic failures of fixed configurations. In the Qwen catastrophe cells we reran, the same $(\text{model}, \text{task}, \text{operator}, \text{timing})$ setting lands within a few points of NoMerge at seeds $1$ and $2$, rather than reproducing the original large regression. Finally, the heatmap should be used for pattern-level conclusions that
aggregate across cells, not for claiming that a specific
$(\text{operator}, \text{timing})$ setting is seed-stably optimal.

\section{Mechanism Diagnostics: Why Do Merge Outcomes Differ?}
\label{sec:diagnostic}
The sweep in \S\ref{sec:sweep} shows that merge can help under the right configuration, but no fixed configuration transfers across tasks. This section asks the next question: what predicts whether a given merge will help? We answer this through three diagnostic probes on the sweep data, each motivating one constraint on the behavioral merge gate in \S\ref{sec:probe}.

\subsection{Prompt-only features cannot predict acceptance}
\label{sec:diagnostic-prediction}

GEPA accepts a merge only if it improves over both parents on a validation subsample, so we ask whether this behavioral decision can be predicted from the prompts alone. If so, the accept gate could be approximated by cheap static features.

For each merge event, we compare the merged child against its two parents and common ancestor using 16 prompt-only features, including lexical overlap, lineage structure, semantic similarity, and LLM-judge scores. These features are intended to capture prompt bloat, parent content loss, semantic novelty, and prompt clarity.

We train logistic classifiers on 213 accepted merges and 52 reconstructed rejected merges, using GroupKFold by sweep cell to prevent leakage across the same $(\text{model}, \text{dataset}, \text{operator}, \text{timing})$ configuration. We evaluate nested prompt-feature sets and include GEPA's behavioral subsample profile as a sanity-check upper bound.

\textbf{Result.} Prompt-only features do not predict GEPA's accept decisions: all prompt-only models achieve ROC-AUC $\leq 0.58$, barely above the random baseline of $0.50$. In contrast, the behavioral subsample-profile feature reaches AUC $=0.987$, because it directly
captures GEPA's accept-gate signal. Thus, accepted and rejected merges are not separable from static prompt properties alone.

\textbf{Implication.} Merge quality is behavioral, not purely textual. A useful merge probe should therefore rely on execution signals such as subsample profiles and plateau behavior, rather than prompt inspection alone. We adopt this constraint in \S\ref{sec:probe}.

\subsection{Reflection dominates merge}
\label{sec:diagnostic-magnitude}

We compare the per-step contributions of reflective mutation and merge. For each candidate in the sweep, we record its origin, its single-step $\Delta\mathrm{val}$ relative to its parent, and the metric-call cost of that step. This gives $n=1{,}300$ candidates for Qwen3-8B and $n=918$ candidates for gpt-4.1-mini across $50$ sweep results per model. Per-task results are shown in Table~\ref{tab:reflect-vs-merge}.

\begin{table}[ht]
\centering
\caption{Mean per-step $\Delta\mathrm{val}$ by operator. Reflection exceeds merge on every task for both models.}
\label{tab:reflect-vs-merge}
\small
\begin{tabular}{@{}lrrrr@{}}
\toprule
& \multicolumn{2}{c}{\textbf{Qwen3-8B}} 
& \multicolumn{2}{c}{\textbf{gpt-4.1-mini}} \\
\cmidrule(lr){2-3}\cmidrule(lr){4-5}
\textbf{Task} & \textbf{Reflect} & \textbf{Merge} 
& \textbf{Reflect} & \textbf{Merge} \\
\midrule
HotpotQA   & $-0.005$ & $-0.008$ & $+0.018$ & $-0.006$ \\
IFBench    & $+0.039$ & $+0.009$ & $+0.012$ & $-0.001$ \\
HoVer      & $+0.020$ & $+0.015$ & $+0.001$ & $-0.004$ \\
MuSiQue    & $+0.018$ & $+0.001$ & $+0.021$ & $-0.005$ \\
2WikiMHQA  & $+0.045$ & $-0.011$ & $+0.040$ & $-0.008$ \\
\bottomrule
\end{tabular}
\end{table}

The same pattern appears on both models. Reflection is where most gains come from, while merge mostly acts as a conservative operator. On Qwen, the average per-step $\Delta\mathrm{val}$ from reflection is about $10\times$ larger than from merge. On GPT, merge is slightly negative on average, while reflection remains positive. This does not imply that merge is useless: rather, merge may be valuable
by recombining partially evolved trajectories and producing candidates that later reflection can refine, even when the merge step itself is not the largest immediate source of gain.

\textbf{Counterfactual.} Reflect-driven prompt growth could be a pathology rather than a value channel. To test this, we reran a fixed GEPA configuration on Qwen3-8B with the reflection LM capped at $2{,}000$ output tokens. Across four Qwen3-8B benchmarks, the cap lowers held-out test performance by an average of $3.46$~pp. The capped run beats NoMerge on only one benchmark and never outperforms the uncapped run.

The failure pattern is clearest on IFBench, the instruction-following benchmark. The $2{,}000$-token cap causes $61$ truncation events, none of which are accepted, effectively starving the reflection channel. This suggests that, at least for instruction-heavy tasks, longer reflective outputs can carry useful optimization signal rather than merely introducing prompt bloat.

\textbf{Implication.}  Merge's direct leverage on mean validation gain is bounded, but this does not make it irrelevant. Its useful role is less to produce the best
candidate immediately, and more to reshape the search trajectory by combining evolved parents, exposing new candidates to later reflection, and controlling tail failures when gated appropriately.

\subsection{Val/test correlation bounds selection reliability}
\label{sec:diagnostic-valtest}

GEPA selects candidates by validation score. If validation correlates poorly with held-out test on a given task, then val-greedy optimization, including any merge policy, has little reliable signal for selecting candidates that improve held-out test performance.

\textbf{Correlation computation.} For each $(\text{model}, \text{task})$ pair, we take the best-validation candidate from each of the $9$ merge configurations and the NoMerge baseline, re-evaluate it on the $300$-example test set, and
compute the Pearson correlation between validation and test scores. This measures whether val-greedy selection transfers to held-out performance.

\begin{table}[t]
\centering
\caption{Per-task validation/test Pearson correlation across the nine merge configurations and NoMerge baseline ($n=10$ per model--task pair).}
\label{tab:val-test-corr}
\small
\begin{tabular}{@{}p{2.0cm}r r@{}}
\toprule
\textbf{Task} & \textbf{Qwen $r$} & \textbf{GPT $r$} \\
\midrule
HotpotQA   & $+0.89$ & $+0.71$ \\
MuSiQue    & $+0.40$ & $+0.83$ \\
IFBench    & $-0.01$ & $+0.00$ \\
HoVer      & $-0.06$ & $+0.63$ \\
2WikiMHQA  & $+0.99$ & $+0.98$ \\
\bottomrule
\end{tabular}
\end{table}

The correlations separate task-level and model-level effects. IFBench has near-zero val/test correlation on both models, suggesting that its validation signal does not reliably support val-greedy optimization. In contrast, HoVer is model-dependent: validation transfers on GPT ($r=+0.63$) but not on Qwen ($r=-0.06$). Thus, when a $(\text{model}, \text{task})$ cell has $r\approx 0$, poor merge performance should be interpreted cautiously, since it may reflect a weak validation-to-test transfer signal rather than a failure of the merge policy itself.

\textbf{Implication.} When val/test correlation is near zero, merge-policy results should be interpreted cautiously: the observed performance may be dominated by validation noise rather than by the merge policy itself. We return to this limitation in \S\ref{sec:discussion}.

\section{Behavioral Probe: Do the Diagnostics Operationalize?}
\label{sec:probe}

\subsection{Probe design}
\label{sec:probe-design}
The diagnostics in \S\ref{sec:diagnostic} guide us to make three predictions for a behavioral merge gate. First, merge decisions should depend on behavioral execution signals rather than prompt-only features
(\S\ref{sec:diagnostic-prediction}). Second, catastrophe prevention should work mainly by abstaining from risky merges, not by repairing them after the fact. This originates from the failure modes in \S\ref{sec:catastrophes} and the limited gain role of merge in \S\ref{sec:diagnostic-magnitude}. Third, the effect of any gate should be model-scale dependent, since val/test correlation varies
across models and tasks (\S\ref{sec:diagnostic-valtest}). We therefore build a behavioral probe from these predictions and ask whether using these signals makes merge more stable: does it avoid
catastrophic cells, abstain on high-risk tasks, and reproduce the model-scale patterns observed in the sweep?

The probe is a four-layer gating policy that intercepts each merge candidate using primarily behavioral execution signals. Each layer maps
to one diagnostic finding; implementation details and thresholds are reported in Appendix~\ref{app:probe-details}.

\begin{itemize}\setlength{\itemsep}{1pt}
\item \textbf{Plateau start gate}: delay merge until reflective search
has consumed enough budget, and
recent validation progress has stalled. This targets the high downside
risk of early merge observed in the sweep.

\item \textbf{Parent-strength and maturity gates}: skip weak parent
pairs and pairs with highly imbalanced lineage evolution across
predictors. These gates target the predictor-maturity imbalance failures.

\item \textbf{Pair selection by complementarity}: prefer parent pairs
with non-overlapping validation errors, so merge is attempted when the
parents appear behaviorally complementary.

\item \textbf{Operator routing}: use the rewriting operator when prompts
are bloated or parent behaviors are highly complementary so as to preserve different trajectories; otherwise use the original merge rule as the conservative default.
\end{itemize}

\begin{figure}[ht]
\centering
\includegraphics[width=\columnwidth]{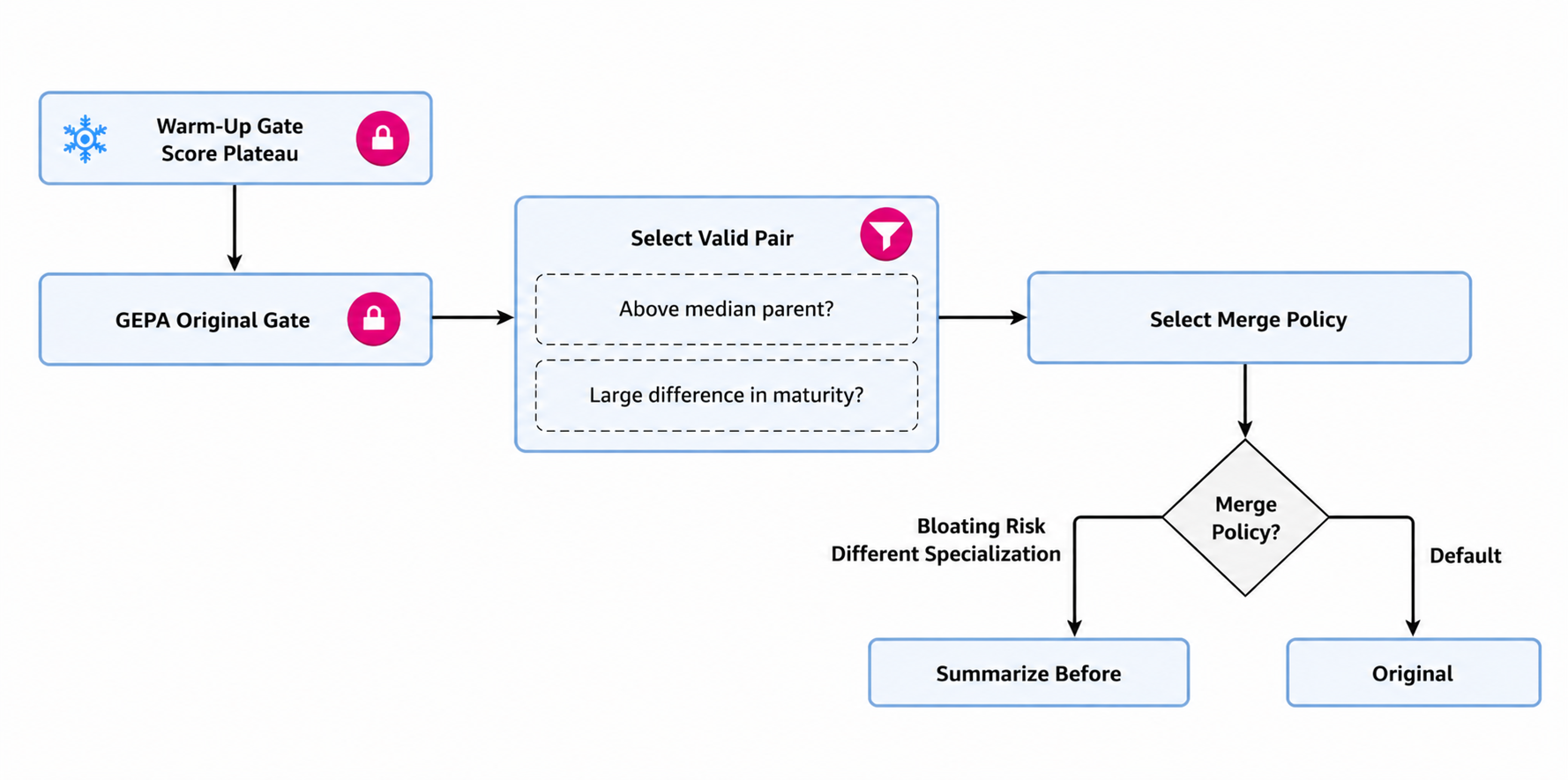}
\caption{Overview of the behavioral probe. Merge is unlocked only after a score plateau and GEPA's original validity checks; parent pairs are filtered by strength and maturity balance, then routed to \textsc{summarize-before} under bloat risk or high complementarity, and to the original rule otherwise.}
\label{fig:probe-pipeline}
\end{figure}

Figure~\ref{fig:probe-pipeline} summarizes the flow. All thresholds are fixed across runs,
and the same probe is evaluated across both models and all tasks.

\subsection{Probe observations}
\label{sec:probe-observations}

\begin{figure}[ht]
\centering
\includegraphics[width=\columnwidth]{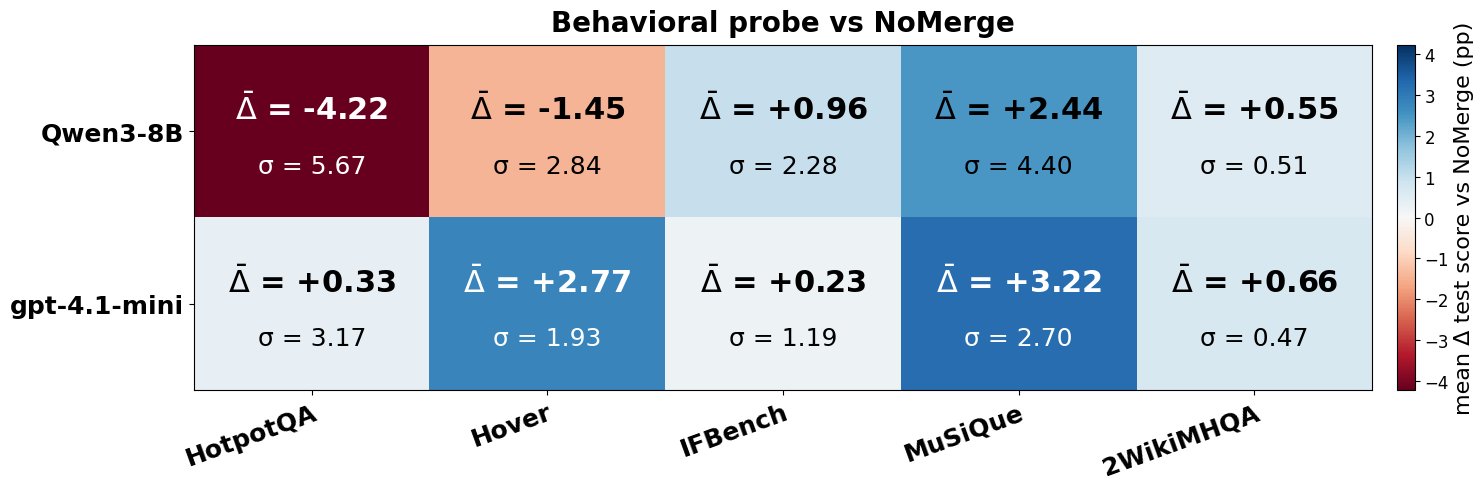}
\caption{\textbf{Behavioral probe vs.\ NoMerge across seeds.}
Each tile reports mean $\bar{\Delta}$ and sample standard deviation $\sigma$.}
\label{fig:probe-heatmap}
\end{figure}

We evaluate the same behavioral probe on both models, using three seeds for four tasks per model: HotpotQA, IFBench, HoVer, and MuSiQue. We additionally evaluate 2WikiMultiHopQA with two GPT seeds and three Qwen seeds to check the documented catastrophe case. Figure~\ref{fig:probe-heatmap} reports the mean and standard deviation relative to the matched NoMerge baseline. Three patterns emerge.

\textbf{Gains track val/test transfer.} The probe is more reliable on gpt-4.1-mini than on Qwen3-8B. On GPT, all five evaluated tasks have positive mean $\Delta$, and HoVer,
MuSiQue, and 2WikiMultiHopQA are positive on every paired seed. On Qwen,
three of the five tasks have positive mean $\Delta$, and only 2WikiMultiHopQA is non-negative on every seed. This matches the val/test analysis in \S\ref{sec:diagnostic-valtest}: when validation performance transfers to test, the behavioral gate can produce stable gains; when it does not, the probe mostly falls within the noise floor.

\textbf{No large tail failures.}
Across 29 paired multi-seed probe results (15 Qwen and 14 GPT, counting only seeds with a matched NoMerge baseline), no run approaches the catastrophic threshold of $-20$~pp; none falls below $-10.00$~pp. The worst case is Qwen $\times$ HotpotQA at seed $1$, which reaches exactly $-10.00$~pp. This is still much smaller than the worst drops in the unconstrained sweep ($-22$ to $-36$~pp; \S\ref{sec:catastrophes}). Thus, the probe does not eliminate variance, but it avoids the large tail failures seen in the sweep. This comparison is not fully controlled, however: the sweep catastrophes come from single-seed runs and did not reproduce under reruns (\S\ref{sec:variance-audit}). A same-seed comparison against the default merge configuration is needed to attribute the smaller tail to the probe rather than to seed variation.

\textbf{Abstention is stochastic, not guaranteed.} On the documented GPT $\times$ 2WikiMultiHopQA catastrophe cell ($-28.66$~pp), the probe accepts no merge events across $1{,}016$ reflective iterations and finishes essentially tied with NoMerge. In the additional
2WikiMultiHopQA reruns, the probe remains close to or above NoMerge: GPT improves across two available seeds, while Qwen remains non-negative across three seeds. Thus, abstention is not a deterministic property of the probe, but its conservative gating reduces exposure to the catastrophe-prone setting and avoids the large tail failure observed in the unconstrained sweep.

Overall, the probe supports the diagnostic story without establishing a deployable policy; we present it as a proof of concept rather than as a recommended replacement for NoMerge. It shows that behavioral signals can reduce large merge failures and are more effective when validation transfers to test, but it does not prove variance reduction over NoMerge in general.

\section{Discussion}
\label{sec:discussion}
\textbf{Merge is not a uniformly improving operator.}
The main takeaway is that GEPA merge is conditionally useful. Reflection provides most of the direct gain in the validation set, while merge plays a more indirect role. It can recombine partially evolved trajectories and change the search path that later reflection continues from. Thus, merge can still matter even when its average per-step gain is small.

\textbf{The instability is mostly contextual.}
The sweep suggests that merge failures are not caused by one inherently broken operator. The same merge mechanism can help or hurt depending on the model, task, timing, and operator variant. Merge is more likely to help when validation transfers to test, parent lineages are mature, and prompt bloat is tolerable. It is more likely to hurt when merge fires too early, parent maturity is imbalanced, or the operator creates overly long prompts.

\textbf{Merge quality is behavioral, not just textual.}
Prompt inspection alone is not enough to decide whether a merge is good. Prompt-only features fail to predict GEPA's accept decisions, while behavioral execution signals recover the accept signal much more directly. Prompt-level features should therefore not be the main basis for accepting a merge. They can flag obvious risks such as prompt bloat, but positive evidence should come from behavioral signals such as parent strength, validation trajectory, complementarity, and val/test transfer.

\textbf{The probe is a safety valve, not an accelerator.}
The role of the behavioral probe is to prevent risky merge behavior, not to guarantee that merge improves every task or always outperforms NoMerge. It reduces large tail failures, but it does not remove seed variance or guarantee improvement over NoMerge on every task. Thus, its main value is downside control rather than uniform gain.

\textbf{Limitations.}
A few limitations remain. First, GEPA has substantial seed-to-seed
variance, sometimes comparable to the differences between merge
configurations. We therefore treat small single-seed gains cautiously and base our conclusions on broader patterns and large tail failures.
Second, we benchmark only two program LMs, gpt-4.1-mini and Qwen3-8B.
Because prompt bloat, reflection quality, and merge tolerance may vary
with model scale, the results may not directly transfer to Qwen3-4B,
Qwen3-14B, or larger proprietary models.
Third, each sweep cell uses a single rollout budget, so we do not know
whether the best timing or operator choice changes under smaller or larger budgets.
Finally, the behavioral probe is only one possible merge gate. It shows that our diagnostics can reduce large failures, but it is not a final or optimal merge policy.

\textbf{Future work.}
Future work should make merge gating and candidate selection more adaptive. First, gate thresholds for parent strength, parent maturity, and prompt bloat should depend on model scale and task feedback granularity rather than being fixed globally. This would test whether smaller models need stricter bloat control, while larger models can tolerate more aggressive recombination.

Second, merge gates should use richer validation trajectories instead of only aggregate scores. For example, tracking whether validation progress has truly plateaued, or whether gains are unstable across minibatches, could help distinguish real stagnation from noisy search variation.

Third, the Pareto frontier itself could include additional evaluation dimensions, such as prompt length, parent complementarity, validation stability, or per-module performance. This may preserve more useful specialist candidates for merge, rather than relying mainly on task score.

Finally, the multi-seed audit should be extended to more seeds, model scales, and budgets. This would test whether val/test correlation, prompt bloat sensitivity, and merge timing preferences change predictably with model size and optimization budget.

\section{Conclusion}
\label{sec:conclusion}
GEPA's merge operator is conditionally useful rather than algorithmically broken. Across five benchmarks and two models, we find that merge can improve performance under appropriate timing and operator choices, but poorly matched configurations can still produce large performance drops. The main mechanism-level lesson is that merge quality is behavioral: prompt features alone do not predict accept decisions, reflection provides most direct validation gains, and weak val/test correlation limits when validation-based selection can help. A behavior-based probe built from these diagnostics reduces large tail failures, but does not eliminate seed variance or establish a final deployment policy. Overall, merge should be treated as a configuration-sensitive trajectory-shaping operator whose value depends on model capacity, parent maturity, task feedback, and behavioral validation signals.

\bibliography{reference}
\bibliographystyle{icml2025}

\newpage
\appendix
\onecolumn
\section{Retrieval and dataset details}
\label{app:retrieval-details}

The five benchmarks split into two retrieval families: BM25-over-corpus (HotpotQA, HoVer) and per-example fixed
context (IFBench, MuSiQue, 2WikiMultiHopQA).
Table~\ref{tab:benchmarks-detail} summarizes each setup. The split is
relevant to merge analysis because BM25-using cells contain an
additional LM module, the query writer, whose instruction is itself
subject to the merge operator, expanding the effective design space. Per-example tasks have a fixed context, so merge only affects
answer-generation instructions.

\begin{table}[ht]
\centering
\caption{Per-benchmark retrieval mode. BM25/Wiki retrieves from a shared Wikipedia abstracts corpus; per-example uses each dataset example's packaged context.}
\label{tab:benchmarks-detail}
\small
\begin{tabular}{@{}p{2.6cm}p{2.0cm}p{2.4cm}@{}}
\toprule
\textbf{Dataset} & \textbf{Retrieval} & \textbf{Context} \\
\midrule
HotpotQA          & BM25 / Wiki  & top-$k$ retrieved \\
HoVer             & BM25 / Wiki  & top-$k$ retrieved \\
IFBench           & none         & instruction only \\
MuSiQue           & per-example  & 20 paragraphs (gold + distractors) \\
2WikiMultiHopQA   & per-example  & 10 paragraphs (gold + distractors) \\
\bottomrule
\end{tabular}
\end{table}

\section{Per-step operator variance}
\label{app:operator-variance}

Table~\ref{tab:reflect-vs-merge-full} reports the full per-step
$\Delta\mathrm{val}$ statistics for reflective mutation and merge,
including standard deviations.

\begin{table}[h]
\centering
\caption{Per-step $\Delta\mathrm{val}$ mean and standard deviation by operator.}
\label{tab:reflect-vs-merge-full}
\small
\begin{tabular}{@{}lrrrr@{}}
\toprule
\textbf{Task} & \textbf{Reflect mean} & \textbf{Merge mean} 
& \textbf{Reflect $\sigma$} & \textbf{Merge $\sigma$} \\
\midrule
\multicolumn{5}{l}{\textit{Qwen3-8B}} \\
HotpotQA   & $-0.005$ & $-0.008$ & $0.047$ & $0.027$ \\
IFBench    & $+0.039$ & $+0.009$ & $0.074$ & $0.023$ \\
HoVer      & $+0.020$ & $+0.015$ & $0.027$ & $0.018$ \\
MuSiQue    & $+0.018$ & $+0.001$ & $0.048$ & $0.015$ \\
2WikiMHQA  & $+0.045$ & $-0.011$ & $0.121$ & $0.022$ \\
\midrule
\multicolumn{5}{l}{\textit{gpt-4.1-mini}} \\
HotpotQA   & $+0.018$ & $-0.006$ & $0.076$ & $0.018$ \\
IFBench    & $+0.012$ & $-0.001$ & $0.022$ & $0.011$ \\
HoVer      & $+0.001$ & $-0.004$ & $0.022$ & $0.029$ \\
MuSiQue    & $+0.021$ & $-0.005$ & $0.069$ & $0.013$ \\
2WikiMHQA  & $+0.040$ & $-0.008$ & $0.103$ & $0.042$ \\
\bottomrule
\end{tabular}
\end{table}

\section{Behavioral Probe Implementation Details}
\label{app:probe-details}

The behavioral probe uses a fixed four-layer gating policy. All
thresholds are fixed across models and tasks.

\paragraph{Plateau start gate.}
Merge is disabled until two conditions hold:
(i) at least $25\%$ of the metric-call budget has been consumed;
(ii) the best tracked score has not improved by more than $\varepsilon=0$ over the last $k=3$ allow-merge calls.

\paragraph{Parent-strength and maturity gates.}
For a candidate parent pair, the parent-strength gate skips the pair if either parent scores strictly below the median score of the active
frontier. The maturity gate computes the combined per-predictor
lineage-mutation counts for the two parents and skips the pair if their
Gini coefficient exceeds $0.50$. This is intended to avoid merging
parents whose improvements are concentrated unevenly across predictors.

A behavioral agreement gate was tested but removed because it was threshold-sensitive and redundant with the complementarity ranker and GEPA's native subsample acceptance check.

\paragraph{Pair selection by complementarity.}
Candidate parent pairs are ranked by behavioral diversity, defined as the fraction of validation examples where exactly one parent is correct. We treat an example as correct if its score is at least $0.5$. If two pairs have the same diversity, we avoid pairs with one weak parent by choosing the pair with the higher minimum parent score. If still tied, we choose the pair with less predictor-token bloat.

\paragraph{Operator routing.}
The probe routes to \textsc{summarize-before} if either condition holds: (i) either parent has a predictor exceeding $L_{\max}=4000$ tokens; or (ii) behavioral diversity exceeds $0.30$. Otherwise, it routes to the original GEPA merge rule.

\section{Per-seed probe results}
\label{app:per-seed-probe}

Per-seed $\Delta$ values for the behavioral probe of
\S\ref{sec:probe-observations}. Summary statistics ($\bar{\Delta}, \sigma$) are reported in Figure~\ref{fig:probe-heatmap}.

\begin{table}[ht]
\centering
\caption{Per-task probe $\Delta$ vs.\ NoMerge across $3$ seeds.}
\label{tab:probe-per-seed}
\small
\begin{tabular}{@{}p{1.5cm}r r r r r@{}}
\toprule
\textbf{Task} & \textbf{$\Delta_{s=0}$} & \textbf{$\Delta_{s=1}$} & \textbf{$\Delta_{s=2}$} & \textbf{$\bar\Delta$} & \textbf{$\sigma$} \\
\midrule
\multicolumn{6}{l}{\textit{gpt-4.1-mini}} \\
HotpotQA   & $+3.66$ & $-2.66$ & $+0.00$ & $+0.33$ & $3.17$ \\
IFBench    & $+1.36$ & $+0.34$ & $-1.02$ & $+0.23$ & $1.19$ \\
HoVer      & $+1.66$ & $+1.66$ & $+5.00$ & $+2.77$ & $1.93$ \\
MuSiQue    & $+1.67$ & $+1.66$ & $+6.34$ & $+3.22$ & $2.70$ \\
2WikiMHQA  & $+1.00$ & $+0.33$ & ---     & $+0.66$ & $0.47$ \\
\midrule
\multicolumn{6}{l}{\textit{Qwen3-8B}} \\
HotpotQA   & $-4.00$ & $-10.00$ & $+1.34$ & $-4.22$ & $5.67$ \\
IFBench    & $+3.57$ & $+0.00$ & $-0.68$ & $+0.96$ & $2.28$ \\
HoVer      & $-4.67$ & $-0.34$ & $+0.67$ & $-1.45$ & $2.84$ \\
MuSiQue    & $+3.34$ & $-2.34$ & $+6.33$ & $+2.44$ & $4.40$ \\
2WikiMHQA  & $+0.00$ & $+0.66$ & $+1.00$ & $+0.55$ & $0.51$ \\
\bottomrule
\end{tabular}
\end{table}

\end{document}